\documentclass[10pt, journal,web]{article}
\usepackage{times}
\usepackage{subfig}
\usepackage[utf8]{inputenc}
\usepackage{utfsym}
\usepackage{multirow}
\usepackage[normalem]{ulem}
\usepackage{amsmath,amssymb,amsfonts}
\usepackage{algorithmic}
\useunder{\uline}{\ul}{}
\usepackage{url}
\def\BibTeX{{\rm B\kern-.05em{\sc i\kern-.025em b}\kern-.08em
    T\kern-.1667em\lower.7ex\hbox{E}\kern-.125emX}}
\usepackage{graphicx}
\usepackage{authblk}

\usepackage{longtable}
\usepackage{pdflscape}
\usepackage[T1]{fontenc}

\usepackage{xcolor}
\usepackage{hyperref}
\usepackage{booktabs}
\hypersetup{
   colorlinks=true,
    linkcolor=blue,
   citecolor=blue,
    urlcolor=blue
}
\usepackage{makecell}
\usepackage{tabularx}

\makeatletter
\renewcommand\AB@affilsepx{, \protect\Affilfont}
\makeatother
\usepackage{tabularx}
\providecommand{\keywords}[1]
{
  \small	
  \textbf{\textit{Keywords---}} #1
}

\begin{document}

\title{\textbf{Edge Artificial Intelligence: A Systematic Review of Evolution, Taxonomic Frameworks, and Future Horizons}}
\author[1, 3, 4]{Mohamad Abou Ali}

\author[1, 2]{Fadi Dornaika\thanks{Corresponding author}}
\affil[1]{\textit{University of the Basque Country}}
\affil[2]{\textit{IKERBASQUE}}
\affil[3]{\textit{Lebanese International University (LIU)}}
\affil[4]{\textit{The International University of Beirut}}

\affil[ ]{

\small\texttt{mohamad.abouali01@liu.edu.lb, fadi.dornaika@ehu.eus}}
\date{}
\maketitle
\begin{abstract}
Edge Artificial Intelligence (Edge AI) embeds intelligence directly into devices at the network edge, enabling real-time processing with improved privacy and reduced latency by processing data close to its source. This review systematically examines the evolution, current landscape, and future directions of Edge AI through a multi-dimensional taxonomy including deployment location, processing capabilities such as TinyML and federated learning, application domains, and hardware types. Following PRISMA guidelines, the analysis traces the field from early content delivery networks and fog computing to modern on-device intelligence. Core enabling technologies such as specialized hardware accelerators, optimized software, and communication protocols are explored. Challenges including resource limitations, security, model management, power consumption, and connectivity are critically assessed. Emerging opportunities in neuromorphic hardware, continual learning algorithms, edge-cloud collaboration, and trustworthiness integration are highlighted, providing a comprehensive framework for researchers and practitioners.

\end{abstract}

 \hspace{10pt}

\section{Introduction}

\subsection{The Imperative for Edge AI: A Paradigm Shift}
The convergence of massive-scale Internet of Things (IoT) deployment and the critical need for real-time, intelligent decision-making has necessitated a fundamental evolution beyond traditional cloud-centric computing architectures. This transition is driven by the impracticalities of cloud-dependent models—namely latency, bandwidth, privacy, and operational resilience—in applications ranging from autonomous vehicles to personalized healthcare. Edge Artificial Intelligence (Edge AI) emerges as the foundational response to these challenges, representing a paradigm that embeds computational intelligence directly into devices at the network periphery \cite{vasuki2024,sipola2022}. By processing data locally, at or near its source, Edge AI enables unprecedented responsiveness, privacy preservation, and operational efficiency \cite{karras2023,sibanda2023}.

\subsection{Methodological Foundation and Analytical Framework}

This review adopts a systematic methodology guided by PRISMA 2020 guidelines \cite{page2021prisma} to ensure a comprehensive, unbiased, and reproducible analysis of the Edge AI landscape. From an initial corpus of over 2,200 identified records, our rigorous screening process yielded 79 primary studies for in-depth qualitative synthesis, forming the analytical core of this review.

\begin{figure}[htbp]
\centering
\includegraphics[width=\textwidth]{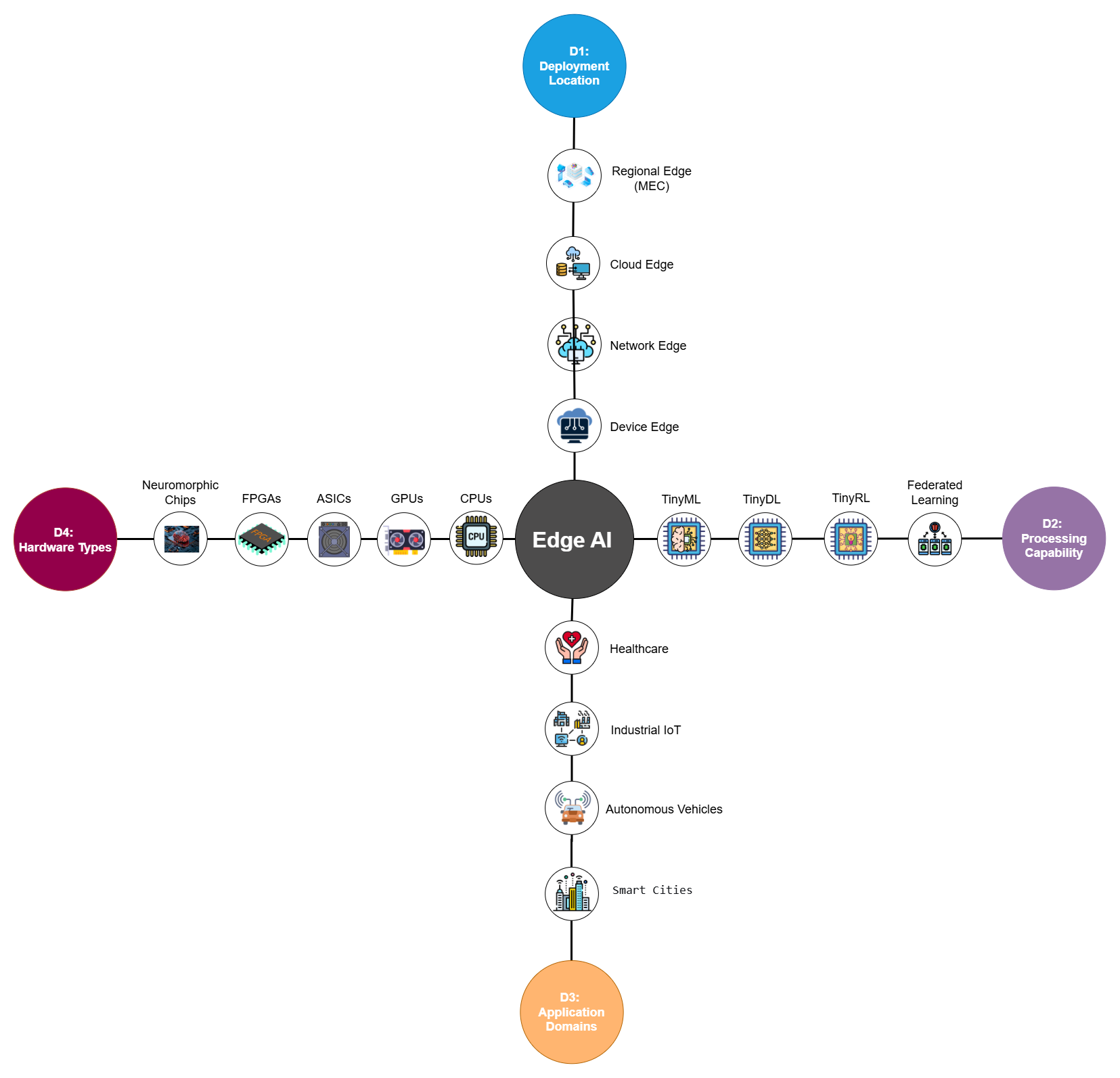}
\caption{Multi-dimensional analytical framework for Edge AI systems, integrating deployment locations, processing capabilities, application domains, and hardware architectures.}
\label{fig:taxonomy}
\end{figure}

The cornerstone of our analysis is a novel multi-dimensional taxonomy (Figure \ref{fig:taxonomy}) that provides an integrated framework for classifying and understanding Edge AI research \cite{Shankar2024}. This taxonomy synthesizes four critical dimensions: deployment location (D1), which spans from \textbf{Device Edge and Network Edge to Regional Edge/Multi-Access Edge Computing (MEC) and Cloud Edge} \cite{ahmed2024, gupta2018}; processing capability (D2), encompassing \textbf{TinyML, TinyDL} \cite{somvanshi2025, lin2023}, \textbf{TinyRL} \cite{wu2024}, \textbf{and federated learning} \cite{wang2019, abreha2022} paradigms; application domain (D3), including \textbf{healthcare} \cite{rocha2024}, \textbf{industrial IoT} \cite{artiushenko2024}, \textbf{autonomous systems} \cite{xie2024}, and \textbf{smart cities} \cite{sharma2024}; and hardware architecture (D4), which covers \textbf{CPUs, ASICs, FPGAs, GPUs, and neuromorphic chips} \cite{alam2024, liang2024, liu2024, bouzidi2022, das2024}. This integrated framework enables the systematic identification of research gaps, technological trade-offs, and future opportunities across the entire Edge AI ecosystem.

\subsection{Research Gaps and Contributions}
Our systematic analysis reveals significant limitations within the current Edge AI literature, which this review addresses through its novel methodological approach. 
First, a notable historical fragmentation exists, as prior surveys \cite{Mendez2022,Singh2023} lack comprehensive historical contextualization by failing to connect modern Edge AI developments to their technological origins in content delivery networks (CDNs) and fog computing. Second, current works exhibit isolated technological analysis; studies such as \cite{Gill2025,Wang2025Optimizing} examine hardware, software, and application layers in isolation, thereby neglecting their critical interdependencies. Third, there is an incomplete challenge assessment across the literature, where existing reviews \cite{Surianarayanan2023,Wang2025Trustworthy} provide only partial coverage of the Edge AI challenge landscape by emphasizing either optimization or security in isolation. 
A systematic comparison of key Edge AI surveys in Table \ref{tab:survey_comparison} further elucidates these research gaps.

\begin{table}[htbp]
\centering
\caption{Comparison of Key Edge AI Surveys}
\label{tab:survey_comparison}
\footnotesize
\begin{tabular}{p{1.5cm}p{2cm}p{3.5cm}p{3.5cm}}
\hline
\textbf{Ref.} & \textbf{Focus} & \textbf{Strengths} & \textbf{Limitations} \\ 
\hline 
\cite{Mendez2022} & Architectures & 
\begin{itemize}
\item Broad coverage
\item HW/SW analysis
\end{itemize} & 
\begin{itemize}
\item Shallow AI depth
\item No benchmarks
\end{itemize} \\ 

\cite{Su2022} & Algorithms & 
\begin{itemize}
\item Technical depth
\item Algorithm comparison
\end{itemize} & 
\begin{itemize}
\item Dense presentation
\item Lacks tools
\end{itemize} \\ 

\cite{Singh2023} & Lightweight AI & 
\begin{itemize}
\item Historical context
\item Application diversity
\end{itemize} & 
\begin{itemize}
\item Edge computing bias
\item Weak metrics
\end{itemize} \\

\cite{Surianarayanan2023} & Optimization & 
\begin{itemize}
\item Multi-layer taxonomy
\item Privacy focus
\end{itemize} & 
\begin{itemize}
\item Conceptual
\item No benchmarks
\end{itemize} \\ 

\cite{Hoffpauir2023} & Edge + LML & 
\begin{itemize}
\item Future insights
\item Trade-off analysis
\end{itemize} & 
\begin{itemize}
\item No validation
\item Vague tools
\end{itemize} \\ 

\cite{Shankar2024} & Technologies & 
\begin{itemize}
\item Case studies
\item Real-time focus
\end{itemize} & 
\begin{itemize}
\item Too brief
\item Weak analysis
\end{itemize} \\

\cite{Meuser2024} & Challenges & 
\begin{itemize}
\item Interdisciplinary
\item Agenda-setting
\end{itemize} & 
\begin{itemize}
\item Abstract
\item No tools
\end{itemize} \\ 

\cite{Gill2025} & Taxonomy & 
\begin{itemize}
\item Robust methodology
\item Collaboration focus
\end{itemize} & 
\begin{itemize}
\item Surface-level
\item No toolchain
\end{itemize} \\ 

\cite{Wang2025Optimizing} & Optimization & 
\begin{itemize}
\item Model compression
\item HW-aware
\end{itemize} & 
\begin{itemize}
\item Dense taxonomy
\item Narrow scope
\end{itemize} \\ 

\cite{Wang2025Empowering} & On-Device AI & 
\begin{itemize}
\item Acceleration focus
\item Foundation models
\end{itemize} & 
\begin{itemize}
\item Scalability gaps
\item Technical overload
\end{itemize} \\ 

\cite{Wang2025Trustworthy} & Trustworthy AI & 
\begin{itemize}
\item Trust framework
\item XAI integration
\end{itemize} & 
\begin{itemize}
\item Few examples
\item Tool gaps
\end{itemize} \\ \hline
\end{tabular}
\normalsize
\end{table}

\subsection{Our Contributions}

This review makes three significant contributions that advance the field of Edge AI. 

First, it provides a novel historical synthesis by tracing the complete evolutionary trajectory from early distributed systems, such as content delivery networks (CDNs), to modern Edge AI paradigms, thereby establishing a critical historical continuity absent from previous surveys. 

Second, it introduces a unified framework through its multi-dimensional taxonomy, which enables an integrated analysis across hardware, software, and application domains to reveal their essential interdependencies and inherent trade-offs. 

Third, it offers a comprehensive challenge analysis that delivers complete coverage of the landscape, including technical constraints, deployment challenges, and fundamental performance trade-offs across the entire Edge AI stack. Together, these contributions provide a foundational and holistic perspective for future research.

\subsection{Paper Organization}
The remainder of this paper is structured as follows: Section \ref{sec:methodology} delineates the PRISMA-guided systematic methodology and multi-dimensional analytical framework. Section \ref{sec:evolution}traces the historical evolution of Edge AI from centralized cloud to distributed intelligence. Section \ref{sec:landscape} presents a taxonomic analysis of the contemporary Edge AI ecosystem. Section \ref{sec:challenges} examines the systemic challenges and fundamental trade-offs. Section \ref{sec:future} projects future research horizons and emerging paradigms. Finally, Section \ref{sec:conclusion} concludes the review by synthesizing key insights and implications.

\section{Research Methodology: A Systematic Multi-Dimensional Review}
\label{sec:methodology}

This review employs a Systematic Literature Review (SLR) methodology, conducted in strict adherence to the Preferred Reporting Items for Systematic Reviews and Meta-Analyses (PRISMA) 2020 guidelines \cite{page2021prisma}. This rigorous approach ensures a transparent, reproducible, and unbiased synthesis of the extant literature on Edge Artificial Intelligence (Edge AI). The process encompassed the formulation of research questions, a comprehensive search strategy, a multi-stage study selection process, systematic data extraction, and analysis through a novel analytical framework.

\subsection{Research Questions}
\label{subsec:rqs}

The review is guided by the following primary Research Questions (RQs), designed to comprehensively map the past, present, and future of Edge AI:

\begin{enumerate}
    \item \textbf{RQ1:} What are the historical milestones and foundational technologies that have shaped the evolution of Edge AI?
    \item \textbf{RQ2:} What constitutes the current state-of-the-art, including core technologies, architectural paradigms, and prevalent types of Edge AI (e.g., TinyML, TinyDL)?
    \item \textbf{RQ3:} What are the significant application domains and their respective impacts?
    \item \textbf{RQ4:} What are the predominant challenges and limitations inherent in Edge AI systems?
    \item \textbf{RQ5:} What are the emerging opportunities and promising future research directions?
\end{enumerate}

These questions provide a structured lens through which the vast body of literature is analyzed and synthesized.

\subsection{Search Strategy}
\label{subsec:search}

A systematic and multi-faceted search strategy was deployed to maximize the retrieval of relevant, high-quality academic literature.

\subsubsection{Keywords and Search Strings}
A comprehensive set of keywords was derived from the research questions and pilot searches to cover the breadth of the domain. The terms included: \textit{"Edge AI"}, \textit{"Edge Artificial Intelligence"}, \textit{"Edge Computing"}, \textit{"TinyML"}, \textit{"Tiny Deep Learning"}, \textit{"Federated Learning at Edge"}, \textit{"On-Device AI"}, and \textit{"Edge Intelligence"}, among others.

These keywords were combined using Boolean operators (AND, OR) to construct complex search queries tailored to each database's syntax. For example: 
\begin{verbatim}
    ("Edge AI" OR "Edge Intelligence") AND ("survey" OR "review")
\end{verbatim}

\subsubsection{Data Sources and Search Period}
The search was executed across eleven leading academic databases and publishers renowned for their coverage of computer science and engineering: IEEE Xplore, ACM Digital Library, SpringerLink, ScienceDirect (Elsevier), arXiv, Wiley Online Library, MDPI, Taylor \& Francis Online, Hindawi, Nature, and Science. An additional 36 records were identified through citation searching of relevant articles.

The initial search yielded 2,220 records. After removing 520 duplicates, a total of 1,700 records were screened by title and abstract. The search period was demarcated from \textbf{January 1, 2000, to June 30, 2025}. This timeframe was selected to capture the foundational work in edge computing and distributed systems, the emergence of key enabling technologies, and the most recent advancements in Edge AI, ensuring a complete historical contextualization.

\subsection{Study Selection and Eligibility Criteria}
\label{subsec:selection}

\begin{figure}[htbp]
\centering
\includegraphics[width=\textwidth]{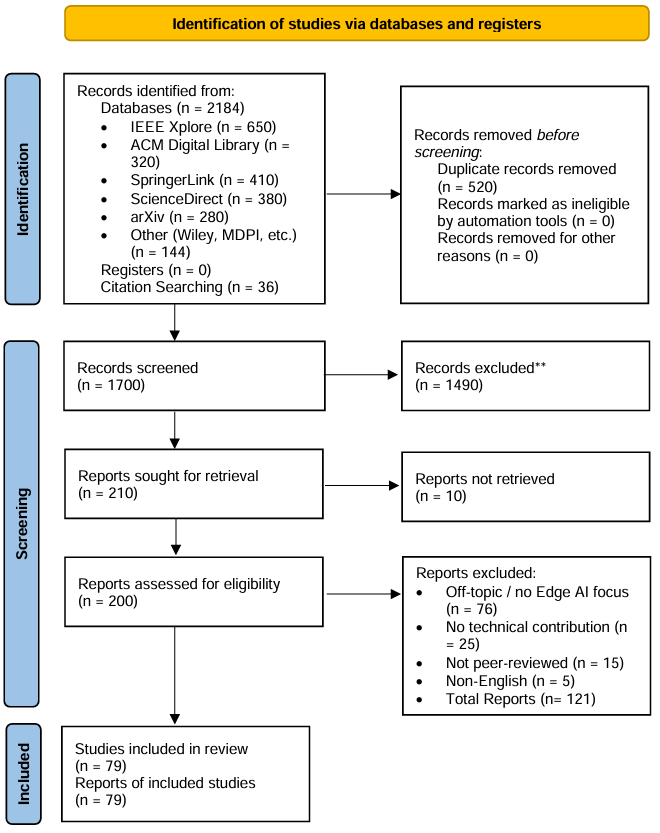}
\caption{PRISMA 2020 flow diagram of the systematic literature identification, screening, and inclusion process.}
\label{fig:prisma_flow}
\end{figure}

The study selection process followed the PRISMA 2020 protocol, as detailed in the flow diagram (Figure \ref{fig:prisma_flow}).

The initial pool of 1,700 records was rigorously screened against formal inclusion and exclusion criteria (Table \ref{tab:inclusion_exclusion}) to ensure both relevance and academic rigor. The title and abstract screening phase resulted in the exclusion of 1,490 records. The full text of the remaining 210 reports was sought for retrieval, of which 10 were not accessible. Consequently, 200 reports were thoroughly assessed for eligibility.

\begin{table}[htbp]
\centering
\caption{Study Inclusion and Exclusion Criteria}
\label{tab:inclusion_exclusion}
\begin{tabularx}{\linewidth}{lX}
\toprule
\textbf{Criterion} & \textbf{Description} \\
\midrule
\textbf{Inclusion} \\
Publication Type & Peer-reviewed journal articles, conference proceedings, and comprehensive survey papers. \\
Language & English. \\
Topic & Primary focus on Edge AI technologies, architectures, applications, challenges, or futures. \\
Time Frame & January 2000 – June 2025. \\
\midrule
\textbf{Exclusion} \\
Publication Type & Short papers (<4 pages), posters, abstracts, editorials, books, theses, and non-academic sources (e.g., blogs, whitepapers). \\
Topic & Focus solely on cloud computing or traditional data centers without an explicit edge component. \\
Accessibility & Full text not retrievable. \\
\bottomrule
\end{tabularx}
\end{table}

Of these, 121 reports were excluded for specific reasons: 76 were off-topic or lacked a primary focus on Edge AI, 25 presented no novel technical contribution, 15 were not peer-reviewed, and 5 were published in a language other than English.

This meticulous process yielded a final corpus of \textbf{79 primary studies} deemed suitable for qualitative synthesis.

\subsection{Data Extraction and Synthesis}
\label{subsec:extraction}

Data from the 79 included studies were extracted into a standardized template. The extracted fields included: bibliographic information (authors, title, year, source), key contributions and findings, identified challenges, and proposed future directions.

\subsection{Analytical Framework: A Multi-Dimensional Taxonomy}
\label{subsec:framework}

To enable a structured and insightful synthesis that moves beyond a descriptive summary, the extracted data was analyzed through a novel multi-dimensional analytical framework, as visualized in Figure \ref{fig:taxonomy}. This framework categorizes each contribution across four interdependent dimensions. 

The first dimension,\textit{ deployment location (D1)}, concerns the physical or logical placement of intelligence, which ranges from the Device Edge — encompassing microcontrollers and sensors—to the Network Edge, including gateways and MEC servers, and finally the Cloud Edge. 

The second dimension, \textit{processing capability (D2)}, identifies the computational paradigm employed, spanning from ultra-constrained TinyML to more capable TinyDL and collaborative Federated Learning. 

The third dimension, \textit{application domain (D3)}, classifies the sector or use-case addressed, such as Healthcare, Industrial IoT, Autonomous Vehicles, or Smart Cities. Finally, the fourth dimension,\textit{ hardware type (D4)}, specifies the underlying processing substrate, which includes ASICs, FPGAs, GPUs, and Neuromorphic Chips.

This integrated framework facilitates a nuanced analysis of trade-offs, synergies, and research gaps across the entire ecosystem. It thereby allows for the identification of under-explored intersections, such as federated learning algorithms optimized for neuromorphic hardware. Consequently, the synthesis presented in subsequent sections is structured to critically examine the literature through this cohesive and original lens.


\section{Historical Evolution: From Cloud to the Intelligent Edge}
\label{sec:evolution}

\subsection{From Cloud to Edge: A Paradigm Shift}
The evolution of computing paradigms represents a continuous quest for optimal efficiency, responsiveness, and scalability, progressively distributing resources closer to the point of demand. This trajectory began with centralized mainframes, evolved through client-server architectures, and culminated in the cloud computing era, which revolutionized data processing through on-demand access to immense, shared computational resources \cite{kiswani2021}. However, the explosive proliferation of connected devices and the emergence of applications requiring real-time processing of massive data volumes at the network periphery exposed critical limitations of the cloud-centric model. Intrinsic bottlenecks related to latency, bandwidth consumption, and data privacy became fundamentally incompatible with the requirements of autonomous systems, real-time analytics, and privacy-sensitive applications \cite{cherukuri2024,charyyev2021}.

\begin{figure}[htbp]
\centering
\includegraphics[width=0.9\textwidth]{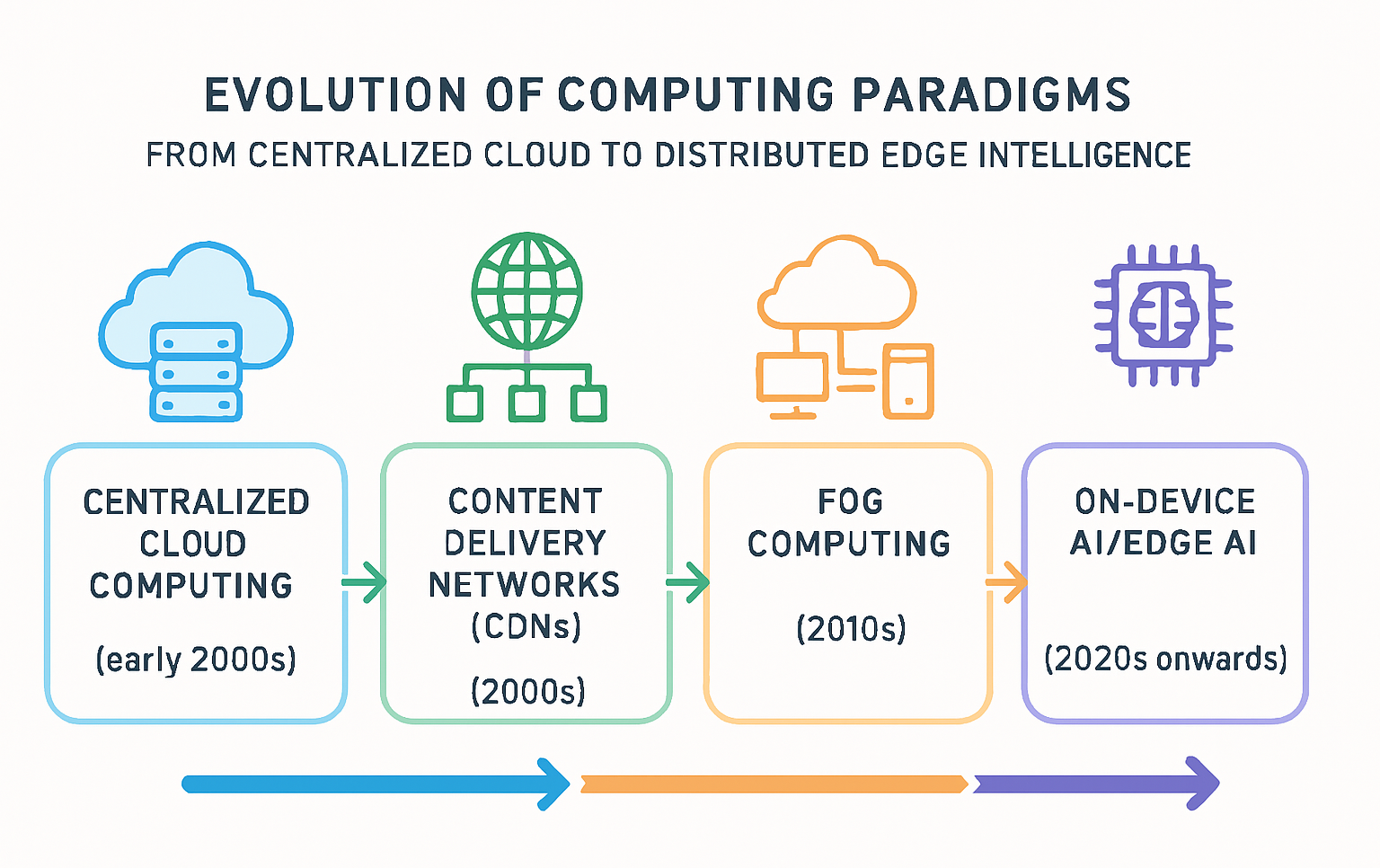}
\caption{Evolution from centralized cloud computing to distributed Edge AI, mapping key technological milestones to the dimensions of our analytical framework.}
\label{fig:evolution}
\end{figure}

This analysis, guided by our methodological framework, reveals that the shift to edge-based processing was not a single event but a series of innovations, each addressing specific dimensions of the cloud limitation problem. As illustrated in Figure \ref{fig:evolution}, this progression established the foundational layers that constitute the modern Edge AI landscape, directly informing the deployment location (D1) and processing capability (D2) dimensions of our taxonomy.

Edge AI, therefore, represents the culmination of this evolutionary trajectory—a deliberate paradigm shift from reliance on distant cloud data centers to the strategic distribution of intelligence adjacent to data sources. This architectural decentralization minimizes long-distance data transmission, thereby critically reducing latency, conserving bandwidth, and enhancing the privacy of sensitive information through localized processing \cite{Avram2014, Sun2020}. The driving force behind this shift is the ubiquitous deployment of smart devices and IoT endpoints that generate continuous data streams in environments where immediate, autonomous decision-making is paramount, such as in industrial automation, autonomous vehicles, and remote healthcare monitoring \cite{choudhary2024,wang2021}.

\subsection{Foundational Technologies: The Pillars of Distributed Intelligence}
The emergence of Edge AI is intrinsically linked to and built upon several foundational technologies that established the core principles of distributed processing. Our systematic review identifies three pivotal technologies that sequentially paved the way for modern Edge AI, each contributing a critical piece to the architectural puzzle.

\subsubsection{Content Delivery Networks (CDNs): The Precursor to Edge Locality}
Content Delivery Networks (CDNs) constituted the earliest widespread form of distributed computing, designed primarily to optimize web content delivery by geographically dispersing cached content closer to end-users \cite{zolfaghari2021}. While their initial focus was content replication rather than computation, CDNs introduced the seminal concept of leveraging network proximity to enhance performance and reduce congestion. This demonstrated the fundamental benefits of localized resource allocation and established the core architectural principle of bringing computation closer to the consumer, laying the essential conceptual groundwork for more sophisticated edge processing paradigms \cite{vagmi2024}. CDNs represent the initial instantiation of what would become the \textbf{Deployment Location (D1)} dimension in our taxonomy.

\subsubsection{Fog Computing: Bridging the Cloud-Edge Divide}
Fog computing emerged as a strategic extension of cloud computing, explicitly designed to bridge the conceptual and architectural gap between centralized cloud resources and edge devices \cite{srirama2023}. By extending cloud services to the network edge, fog computing enabled computation, storage, and networking capabilities to be performed in closer proximity to data sources. Fog nodes—often implemented on routers, switches, or dedicated servers—functioned as intelligent intermediaries between edge devices and the cloud, providing crucial localized processing and reducing the dependency on continuous, high-bandwidth cloud connectivity \cite{gupta2018,han2021}. This architecture introduced a hierarchical computing model, a concept central to modern Edge AI, which explicitly defined different processing tiers at varying distances from the data source. Fog computing directly informs the hierarchical nature of both the \textbf{Deployment Location (D1)} and \textbf{Processing Capability (D2)} dimensions in our taxonomy.

\subsubsection{Mobile Edge Computing (MEC): The Ultralow-Latency Enabler}
Mobile Edge Computing (MEC), standardized as Multi-access Edge Computing, advanced the distribution paradigm by integrating cloud computing capabilities directly within the radio access network (RAN) infrastructure \cite{ahmed2024}. By positioning computation and storage resources at the base station level, MEC brings unprecedented proximity to mobile users and devices, offering ultralow latency and high bandwidth essential for advanced applications. This proximity is critical for latency-sensitive use cases such as augmented reality, virtual reality, and real-time video analytics, where millisecond-scale delays significantly impact user experience and system performance \cite{khan2022,li2025}. MEC platforms enable application deployment at cellular base stations or access points, facilitating immediate data processing and responses, a capability that defines the high-performance end of the \textbf{Processing Capability (D2)} dimension for mobile scenarios.

\subsection{The Rise of On-Device AI: The Ultimate Realization of Edge Intelligence}
The convergence of these foundational technologies, coupled with breakthroughs in hardware miniaturization and energy-efficient AI algorithms, has catalyzed the most significant evolution: the rise of On-Device AI. This paradigm refers to the execution of sophisticated AI models directly on end-user devices—such as smartphones, wearables, sensors, and microcontrollers—often independent of continuous cloud or fog connectivity \cite{moon2022,wang2025}.

On-Device AI represents the ultimate expression of edge intelligence, offering transformative advantages in privacy, latency, and operational autonomy. By ensuring data never leaves the device, it fundamentally mitigates privacy and security risks associated with data transmission. The elimination of network latency ensures genuine real-time inference, and devices maintain intelligent functionality even in offline or intermittently connected environments \cite{ubale2025,xu2025}. This shift has precipitated remarkable innovation in highly optimized, compact AI models, spurring not only TinyML but also the more capable Tiny Deep Learning (TinyDL) and Tiny Reinforcement Learning (TinyRL) paradigms, alongside the development of specialized hardware accelerators tailored for severely resource-constrained environments. 

This historical analysis, structured through our methodological framework, demonstrates that the evolution of Edge AI has been a process of continuous refinement across the dimensions of deployment, processing, and hardware. The following section will delve into the current landscape of Edge AI, utilizing our taxonomy to provide a structured analysis of the technologies and architectures that have emerged from this evolutionary journey.


\section{A Taxonomic Analysis of the Contemporary Edge AI Ecosystem}
\label{sec:landscape}

The contemporary state of Edge AI is defined by a complex, synergistic ecosystem of specialized hardware, optimized software stacks, and efficient communication protocols, all orchestrated to enable intelligent processing under stringent resource constraints. This section employs the multi-dimensional taxonomy introduced in Figure \ref{fig:taxonomy} to provide a structured analysis of this landscape. We deconstruct the ecosystem into its core technological pillars and then examine the dominant paradigms (TinyML, TinyDL, Federated Learning) that emerge from the interplay of these pillars, concluding with a analysis of their application across critical domains.

\subsection{The Core Technology Stack: Hardware, Software, and Communication}
\label{subsec:technologies}
The effective deployment of Edge AI is predicated on the co-design of hardware, software, and communication layers. This tripartite foundation directly maps to the \textbf{Hardware Type (D4}) and \textbf{Processing Capability (D2)} dimensions of our taxonomy, enabling the diverse \textbf{Deployment Locations (D1)} discussed in Section 3.

\begin{figure}[htbp]
\centering
\includegraphics[width=0.8\textwidth]{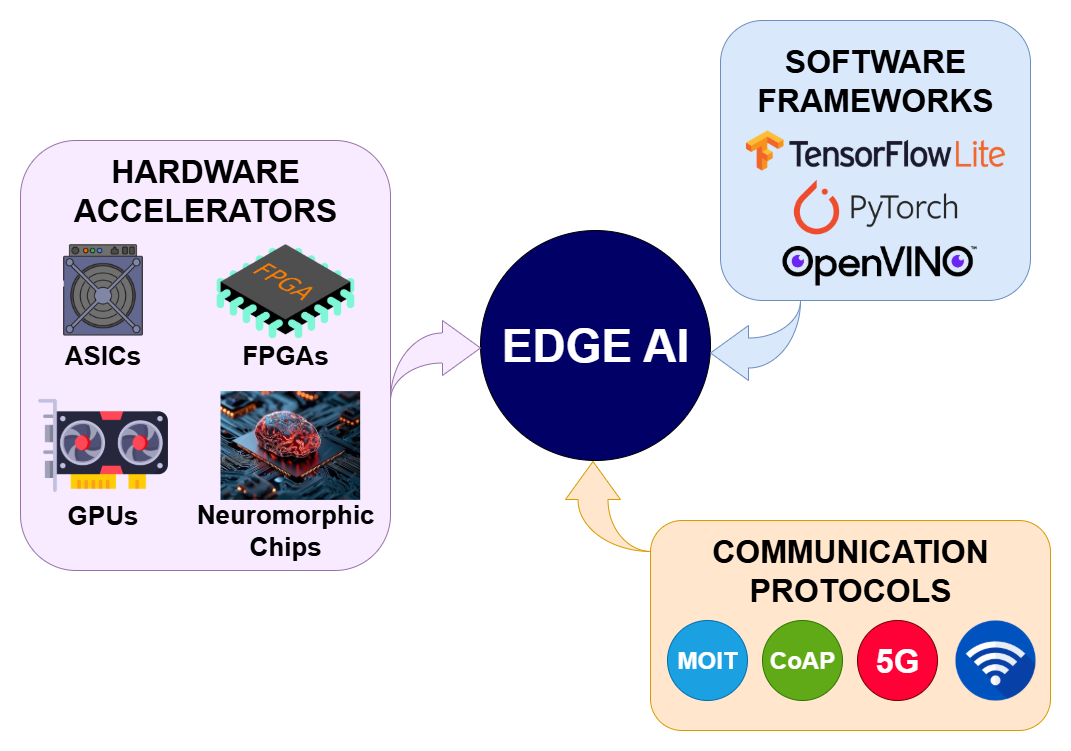}
\caption{The Edge AI technology stack, illustrating the synergistic relationship between hardware accelerators, software frameworks, and communication protocols that enable intelligent processing across the deployment continuum.}
\label{fig:technologies}
\end{figure}

The Edge AI technology stack (Figure \ref{fig:technologies}) is not a collection of isolated components but a tightly integrated hierarchy. breakthroughs in hardware accelerators (e.g., Google's Edge TPU) create the substrate for efficient inference, which software frameworks (e.g., PyTorch Mobile) leverage through advanced optimization techniques, while communication protocols (e.g., 5G) enable the orchestration of intelligence across the edge-to-cloud continuum.

\subsubsection{Hardware Accelerators for Edge AI: The Physical Substrate (D4)}

The computational exigencies of contemporary artificial intelligence models, particularly deep neural networks, substantially surpass the capabilities of conventional general-purpose processors in resource-constrained edge environments. This discrepancy has precipitated the development of specialized hardware accelerators, each embodying distinct architectural trade-offs within the \textbf{Hardware Type (D4)} dimension of our taxonomic framework. These co-processors are fundamentally engineered to maximize computational efficiency, quantified as trillions of operations per second per watt (TOPS/W) \cite{alam2024,liang2024}.

Four principal architectural paradigms dominate this landscape. \textbf{Application Specific Integrated Circuits (ASICs)}, epitomized by Google's Edge TPU, represent the zenith of performance and energy efficiency for fixed-function, high-volume inference workloads; however, this optimization comes at the expense of architectural flexibility due to their hardened circuitry. \textbf{Field-Programmable Gate Arrays (FPGAs)}, such as the Xilinx Versal series, occupy a middle ground by offering reconfigurable logic fabrics that balance computational efficiency with post-deployment adaptability, making them particularly suitable for evolving algorithmic requirements and prototyping applications \cite{liu2024}. \textbf{Graphics Processing Units (GPUs)}, including power-optimized variants like the NVIDIA Jetson platform, leverage massive parallel processing capabilities to accelerate complex computational workloads such as real-time video analytics, though this often necessitates accepting higher power envelopes \cite{bouzidi2022}. Finally, \textbf{neuromorphic computing platforms}, exemplified by Intel's Loihi architecture, constitute a paradigm-shifting approach that emulates biological neural networks through event-driven, asynchronous processing, thereby offering transformative potential for ultra-low-power operation on sparse, temporal data streams \cite{das2024}.

This heterogeneous ecosystem of accelerator architectures demonstrates that no single solution optimally addresses all edge computing constraints, thereby necessitating careful architectural co-design across the hardware-software stack to meet specific application requirements within the Edge AI domain.

\begin{table}[htbp]
\centering
\caption{Performance characteristics and trade-offs of dominant Edge AI hardware platforms \cite{alam2024, liu2024, das2024}, cataloged by the Hardware Type (D4) dimension.}
\begin{tabularx}{\textwidth}{lXXXX}
\toprule
\textbf{Type (D4)} & \textbf{Example} & \textbf{Pros} & \textbf{Cons} & \textbf{Optimal Deployment (D1) \& Use Case} \\
\midrule
\textbf{ASICs} & Google Edge TPU & High TOPS/W, low latency & Fixed architecture, costly design & Network/Cloud Edge; Fixed, high-volume inference \\
\textbf{FPGAs} & Xilinx Versal & Reconfigurable, energy-efficient & High development complexity & Network Edge; Evolving models, prototyping \\
\textbf{GPUs} & NVIDIA Jetson & High parallelism, flexible & Power-hungry, expensive & Device/Network Edge; Video analytics, training \\
\textbf{Neuromorphic} & Intel Loihi & Event-driven, ultra-low power & Niche programming model & Device Edge; Sensor fusion, sparse data \\
\bottomrule
\end{tabularx}
\label{tab:hardware}
\end{table}

Selecting the appropriate accelerator (Table \ref{tab:hardware}) is a critical system-level decision dictated by the constraints of the \textbf{Application Domain (D3)} and the target \textbf{Deployment Location (D1)}. ASICs deliver unmatched efficiency for static workloads at the network edge, while FPGAs provide crucial adaptability for industrial settings. Neuromorphic chips, though nascent, offer a disruptive potential for next-generation always-on sensing applications at the extreme device edge.

\subsubsection{Edge AI Frameworks and Runtimes: The Software Abstraction Layer}

To effectively harness the computational capabilities of heterogeneous edge hardware, a sophisticated suite of software frameworks has emerged, specializing in model optimization, quantization, and efficient inference execution. These frameworks constitute the critical software abstraction layer that operationalizes the \textbf{Processing Capability (D2)} dimension for any given \textbf{Hardware Type (D4)} within our taxonomy \cite{rexha2021,dagli2021}.

Several prominent frameworks exemplify this technological stratum. \textit{\textbf{TensorFlow Lite and PyTorch Mobile}} provide lightweight runtimes derived from their respective mainstream machine learning ecosystems, specifically engineered for deployment on mobile and embedded devices with stringent requirements for low latency and minimal binary footprint. The \textit{\textbf{OpenVINO Toolkit}} represents a vendor-specific optimization suite that enhances deep learning model performance across Intel's heterogeneous hardware portfolio, including CPUs, GPUs, FPGAs, and Vision Processing Units (VPUs), thereby maximizing inference efficiency on dedicated silicon architectures. In contrast, O\textbf{\textit{NNX Runtime}} embodies a cross-platform inference paradigm that promotes framework interoperability by enabling models trained within one ecosystem (e.g., PyTorch) to be deployed through hardware-optimized runtimes from alternative toolchains, effectively mitigating vendor lock-in concerns.

Collectively, these frameworks provide crucial abstraction of underlying hardware complexity, enabling developers to concentrate on algorithmic innovation and model design while the software stack manages the intricate challenges of executing computational graphs efficiently across diverse accelerator architectures. This architectural separation between hardware capabilities and software implementation fundamentally enables the practical deployment of advanced AI models within the constrained environments characteristic of edge computing scenarios.

\subsubsection{Communication Protocols for Edge AI: The Connectivity Fabric}
Efficient and reliable data exchange is the connective tissue of distributed Edge AI systems, enabling collaborations between devices at different \textbf{Deployment Locations (D1)}. The choice of protocol is dictated by the constraints of the \textbf{Application Domain (D3)}, particularly its latency and bandwidth requirements \cite{shi2020,mwase2022}.

\begin{itemize}
\item \textbf{MQTT \& CoAP:} Lightweight messaging protocols designed for constrained devices and unreliable networks, ideal for telemetry data transmission in IoT scenarios.
\item \textbf{5G and Beyond:} The ultra-reliable low-latency communication (URLLC) and enhanced mobile broadband (eMBB) capabilities of 5G are transformative enablers, facilitating real-time collaboration between edge devices and servers for advanced applications like autonomous driving and augmented reality \cite{kartsakli2023}.
\end{itemize}

\subsection{Paradigms of Processing: From TinyML to Federated Learning}
\begin{figure}[htbp]
\centering
\includegraphics[width=\textwidth]{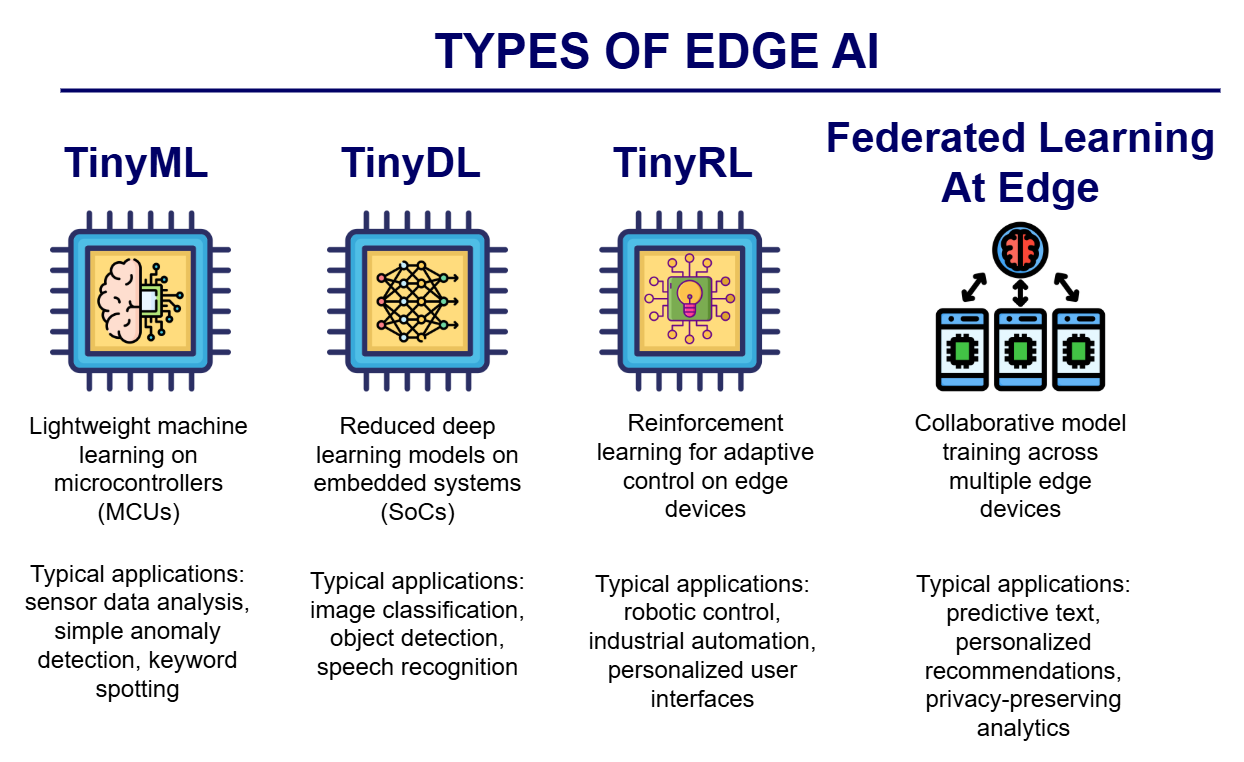}
\caption{The spectrum of Edge AI paradigms, classified by Processing Capability (D2), ranging from ultra-constrained TinyML to collaborative Federated Learning.}
\label{fig:types}
\end{figure}

Edge AI is not monolithic but comprises a spectrum of paradigms tailored to specific resource constraints and processing requirements. These paradigms represent different points along the \textbf{Processing Capability (D2)} dimension, as visualized in Figure \ref{fig:types}.

The strategic selection of a paradigm involves navigating a complex trade-off space, as systematically compared in Table \ref{tab:paradigms}. This decision is primarily driven by the constraints of the \textbf{Hardware Type (D4)} and the requirements of the \textbf{Application Domain (D3)}.

\begin{landscape}
\begin{table}[htbp]
\caption{Comparative Analysis of Edge AI Paradigms along the Processing Capability (D2) dimension.}
\label{tab:paradigms}
\begin{tabular}{p{4cm} p{4.2cm} p{3.9cm} p{3.9cm} p{5.5cm}} 
\toprule
\textbf{Feature} & \textbf{TinyML} & \textbf{TinyDL} & \textbf{TinyRL} & \textbf{Federated Learning at the Edge} \\
\midrule
Target Hardware (D4) & Microcontrollers & Embedded SoCs & Embedded SoCs & Heterogeneous Device Fleets \\
Compute Resources & KBs of RAM, µW–mW & MBs of RAM, mW–W & MBs of RAM, mW–W & Aggregated resources \\
Typical Applications (D3) & Anomaly detection & Object detection & Robotic control, & Privacy-preserving analytics \\
Key Advantage & Ultra-low power & Efficient deep learning & Adaptive decision-making & Privacy, leverage distributed data \\
Key Challenge & Extreme constraints & Model optimization & Sample efficiency, policy compression & Communication overhead, heterogeneity \\
\bottomrule
\end{tabular}
\end{table}
\end{landscape}

\subsubsection{TinyML: Intelligence at the Extremes (D2)}
TinyML operates at the most constrained end of the \textbf{Processing Capability (D2)} spectrum, deploying highly optimized models onto microcontrollers with kilobytes of memory and microwatt power budgets \cite{immonen2022,lin2023}. Its value proposition is enabling always-on, always-sensing capabilities for applications like industrial monitoring and wearable health devices, where data privacy and power autonomy are paramount \cite{ray2022,han2022}. The core innovation lies in extreme model compression and quantization techniques that strip neural networks down to their bare essentials without sacrificing critical functionality.

\subsubsection{TinyDL: Embedded Deep Learning (D2)}
TinyDL occupies the middle ground, enabling more complex deep learning tasks (e.g., image classification, object detection) on embedded systems like Raspberry Pi or NVIDIA Jetson platforms \cite{somvanshi2025,roveri2023}. These devices possess marginally more resources (MBs of RAM, watt-level power), allowing for the execution of deeper networks. The focus shifts from extreme compression to sophisticated optimization techniques like pruning, knowledge distillation, and neural architecture search to balance accuracy, latency, and power consumption for applications in smart cameras, drones, and robotics \cite{darvish2017,burrello2024}.

\subsubsection{TinyRL: Reinforcement Learning at the Edge (D2)}
Tiny Reinforcement Learning (TinyRL) represents the cutting edge of on-device learning, enabling autonomous decision-making and control policies directly on resource-constrained hardware \cite{wu2024}. Unlike the supervised learning focus of TinyML and TinyDL, TinyRL algorithms learn optimal behaviors through interaction with their environment, making them ideal for applications in robotics, industrial control, and network management where systems must adapt in real-time without cloud dependency \cite{du2023}. The primary challenge lies in compressing the high sample complexity and memory requirements of traditional RL into the extreme constraints of the device edge, often leveraging techniques like policy distillation and efficient experience replay.

\subsubsection{Federated Learning: Collaborative Intelligence}
Federated Learning (FL) is a unique paradigm that transcends a single \textbf{Deployment Location (D1)}. It is a privacy-preserving distributed training method that leverages the collective computational power of a heterogeneous fleet of edge devices (\textbf{D4}) \cite{thomas2025,abreha2022}. Instead of centralizing raw data, FL trains models locally on each device and aggregates only model updates. This makes it particularly suited for \textbf{Application Domains (D3)} with stringent privacy concerns, such as healthcare and finance, though it introduces challenges in communication efficiency and handling device heterogeneity \cite{brecko2022,hemmati2025}.

\subsubsection{Other Variants: Deployment Location (D1) Specialization}

Beyond the core computational paradigms previously examined, the Edge AI ecosystem exhibits specialized architectural implementations distinguished primarily by their positioning within the \textbf{Deployment Location (D1)} dimension. Two particularly significant specializations merit explicit consideration. 

First, \textbf{Edge AI on Gateways/Servers} (including Regional Edge/MEC nodes) represents an approach that deploys more capable models (TinyDL, TinyRL) on robust appliances, enabling sophisticated processing like real-time multi-sensor fusion in smart manufacturing and urban infrastructure \cite{wang2025a,nguyen2024}.

Second,\textbf{ Edge AI in Mobile Device}s constitutes a distinct paradigm that harnesses dedicated neural processing units (NPUs) integrated within smartphones to facilitate advanced on-device applications including augmented reality interfaces and computational photography enhancements \cite{hirsch2025,wang2019}.

These deployment-specific implementations demonstrate that the physical and logical placement of computational resources serves as a fundamental architectural determinant that directly governs system capabilities, operational constraints, and performance characteristics across diverse Edge AI applications. The strategic selection of deployment location thus represents a critical design consideration that profoundly influences both the technical feasibility and practical efficacy of Edge AI solutions within their intended operational environments.

\begin{table}[htbp]
\centering
\caption{Deployment Tiers of Edge AI, synthesizing Hardware Type (D4), resource constraints, and typical Application Domains (D3).}
\begin{tabularx}{\textwidth}{p{2cm}p{2.5cm}p{1.3cm}p{1.3cm}p{3.5cm}}
\toprule
\textbf{Category} & \textbf{Device Examples (D4)} & \textbf{Memory} & \textbf{Power} & \textbf{Typical Use Cases (D3)} \\
\midrule
\textbf{TinyML} & Micro-controllers & KBs & µW--mW & Keyword spotting, vibration monitoring \\
\textbf{TinyDL} & Embedded SoCs & MBs & mW--W & Real-time object detection, drones \\
\textbf{Edge Servers} & Gateways, MEC nodes & GBs & W--kW & Smart city analytics, multi-sensor fusion \\
\bottomrule
\end{tabularx}
\label{tab:deployment}
\end{table}

The stratification of the Edge AI landscape into distinct deployment tiers (Table \ref{tab:deployment}) is a direct consequence of the trade-offs analyzed through our taxonomic framework. The choice of tier dictates the feasible paradigms and ultimately the applications that can be successfully deployed.

\subsection{Application Domains: Transformative Impact Across Industries}
Edge AI is fundamentally transforming industries by enabling intelligent, real-time decision-making at the data source. Its value proposition—local processing, reduced latency, and enhanced privacy—makes it indispensable across diverse sectors. The following analysis, structured by the \textbf{Application Domain (D3)} dimension, highlights the most impactful use cases, with their benefits summarized in Table \ref{tab:applications}.

\begin{table}[htbp]
\centering
\caption{Edge AI Applications categorized by Application Domain (D3), highlighting domain-specific benefits.}
\begin{tabularx}{\textwidth}{lXX}
\toprule
\textbf{Domain (D3)} & \textbf{Application Examples} & \textbf{Key Benefits of Edge AI} \\
\midrule
Smart Cities & Traffic management, public safety & Real-time processing, low-latency decision-making \\
Industrial IoT & Predictive maintenance, quality control & On-device analytics, minimized downtime \\
Autonomous Vehicles & Object detection, navigation & Ultra-low latency, enhanced safety, independence \\
Healthcare & Remote patient monitoring, diagnostics & Strong data privacy, real-time insights \\
Retail & Inventory tracking, customer analytics & Real-time insights, improved user experience \\
\bottomrule
\end{tabularx}
\label{tab:applications}
\end{table}

\subsubsection{Smart Homes and Cities}
In smart homes, Edge AI enhances privacy and responsiveness. Devices like smart speakers, security cameras, and thermostats can process voice commands, detect intruders, or optimize energy consumption locally without sending sensitive data to the cloud \cite{thakur2024,sheikh2025}. This on-device processing ensures immediate responses and reduces concerns about data breaches. For instance, a smart doorbell with Edge AI can identify known visitors or detect suspicious activity in real-time, sending alerts only when necessary \cite{patel2020}.

In smart cities, Edge AI plays a crucial role in managing urban infrastructure and services. Applications include intelligent traffic management systems that optimize signal timings based on real-time traffic flow detected by edge cameras \cite{sharma2024}, smart streetlights that adjust illumination based on pedestrian and vehicle presence \cite{omar2022}, and waste management systems that optimize collection routes using AI-powered sensors on bins \cite{choubey2024}. These deployments leverage Edge AI to improve efficiency, sustainability, and public safety.

\subsubsection{Industrial IoT (IIoT)}
Edge AI is a cornerstone of Industry 4.0, enabling predictive maintenance, quality control, and operational optimization in manufacturing and industrial settings. By deploying AI models directly on factory equipment, sensors can monitor machine health, detect anomalies, and predict potential failures before they occur, minimizing downtime and reducing maintenance costs \cite{argungu2023,artiushenko2024}. For example, vibration sensors with embedded AI can analyze machine vibrations to identify early signs of wear and tear, triggering alerts for proactive maintenance \cite{ji2025}.

Furthermore, Edge AI facilitates real-time quality inspection on production lines, where cameras with embedded AI can identify defects in products with high accuracy and speed, ensuring consistent product quality \cite{mandapaka2023}. This localized processing is critical in environments where network latency to the cloud could lead to significant production delays or errors.

\subsubsection{Autonomous Vehicles}
Autonomous vehicles are perhaps one of the most demanding applications for Edge AI, requiring ultra-low latency and highly reliable real-time decision-making. Self-driving cars must process vast amounts of sensor data (from cameras, LiDAR, radar, etc.) instantaneously to perceive their environment, predict the behavior of other road users, and make critical navigation decisions \cite{rahmati2025,manivannan2025}. Cloud-based processing for such tasks is impractical due to the inherent latency, which could lead to catastrophic delays. Edge AI enables these vehicles to operate autonomously and safely by performing complex AI computations directly on board \cite{xie2024}. This includes object detection, lane keeping assistance, pedestrian recognition, and real-time path planning, all executed at the edge to ensure immediate responses to dynamic road conditions \cite{ahmed2025}.

\subsubsection{Healthcare}
In healthcare, Edge AI offers transformative potential, particularly in remote patient monitoring, diagnostics, and personalized medicine. Wearable health devices equipped with Edge AI can continuously monitor vital signs, detect anomalies, and alert patients or healthcare providers to critical changes, all while preserving patient privacy by processing sensitive data on-device \cite{sathiya2025,rocha2024}. For instance, an Edge AI-powered ECG device can detect arrhythmias in real-time, providing immediate feedback to the user or triggering emergency services \cite{huang2024}.

Edge AI also supports intelligent medical imaging analysis at the point of care, allowing for faster preliminary diagnoses in remote clinics or emergency settings without relying on cloud connectivity \cite{xu2025a}. This can significantly reduce the time to diagnosis and improve patient outcomes, especially in underserved areas.

\subsubsection{Retail}
Edge AI is revolutionizing the retail sector by enhancing customer experience, optimizing store operations, and improving inventory management. In smart retail environments, Edge AI-powered cameras can analyze customer traffic patterns, optimize product placement, and detect shoplifting in real-time, without transmitting continuous video feeds to the cloud \cite{biswas2021,rashvand2025}. This not only improves security but also provides valuable insights into customer behavior while maintaining privacy.

Furthermore, Edge AI can facilitate automated checkout systems, smart shelves that monitor inventory levels, and personalized advertising displays that adapt to customer demographics or preferences, all processed locally to ensure immediate and relevant interactions \cite{savit2023,islam2024}. These applications demonstrate how Edge AI can create more efficient, secure, and customer-centric retail experiences.


\section{Systemic Challenges and Fundamental Trade-Offs}
\label{sec:challenges}

\begin{figure}[htbp]
\centering
\includegraphics[width=\textwidth]{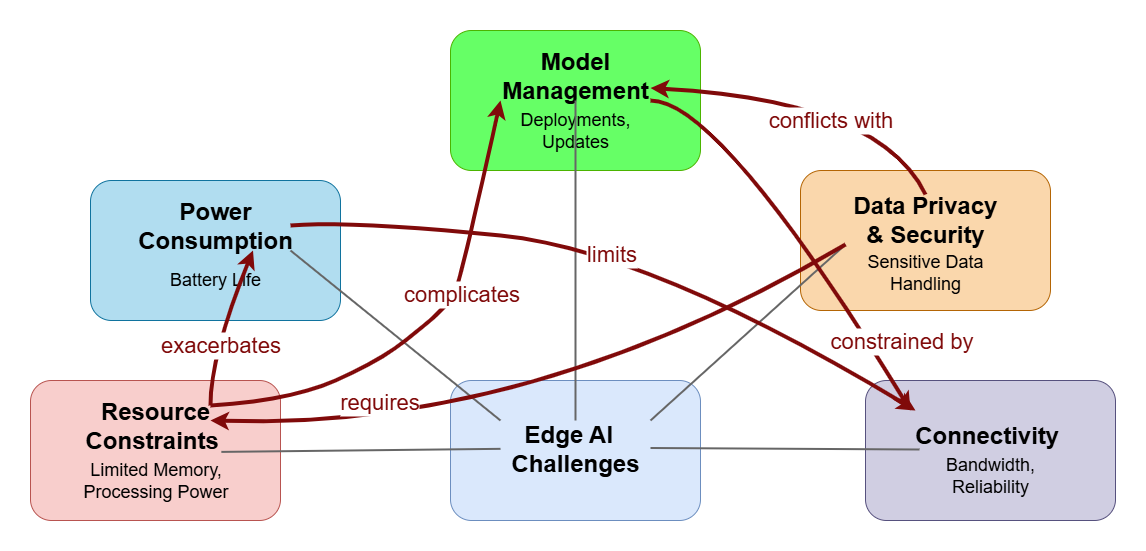}
\caption{Systemic challenges in Edge AI, depicting five core challenges (resource constraints, data privacy \& security, model management, power consumption, and connectivity) and their causal interrelationships (e.g., "exacerbates", "requires") across the technology stack.}
\label{fig:challenges}
\end{figure}

Despite its transformative potential, the widespread deployment of Edge AI is contingent upon overcoming significant and interconnected challenges. These limitations are not merely technical hurdles but fundamental trade-offs inherent to the distributed, resource-constrained, and heterogeneous nature of edge environments. As illustrated in Figure \ref{fig:challenges}, these five core challenges—resource constraints, data privacy, model management, power consumption, and connectivity—and their interconnections span the entire stack, from hardware and algorithms to security and systems management. This section analyzes these barriers through the lens of our multi-dimensional taxonomy, examining how constraints in one dimension (e.g., Hardware Type - D4) precipitate challenges in another (e.g., Processing Capability - D2 or Deployment - D1).

\subsection{Resource Constraints: The Fundamental Trade-Off}
The most fundamental challenge in Edge AI stems from the severe resource constraints intrinsic to devices at the \textbf{Device Edge (D1)}, particularly those running \textbf{TinyML (D2)}. Furthermore, emerging paradigms like TinyRL introduce additional complexity, as their training-inference cycles and memory requirements for experience replay present novel optimization hurdles beyond those of static inference or even federated learning. These constraints—encompassing computational power, memory (RAM and storage), and energy budgets—define the design space and force a continuous trade-off between model accuracy, latency, inference speed, and power consumption \cite{dong2022,jia2023}.

While techniques like model compression (pruning, quantization), efficient neural architectures (e.g., MobileNets), and specialized \textbf{hardware accelerators (D4)} are crucial mitigations, they are not panaceas. For instance, aggressive quantization can achieve 3--5× energy savings but often at the cost of non-negligible accuracy loss, especially for complex models \cite{husom2025,muhoza2023}. Consequently, developing robust AI capabilities for devices with kilobyte-scale memory and milliwatt power budgets remains a formidable engineering challenge that dictates the feasible \textbf{Application Domains (D3)} for a given hardware class.

\subsection{Data Privacy and Security: The Expanded Attack Surface}
\label{sec:privacy}

While Edge AI inherently enhances privacy by processing data locally, reducing its exposure over networks, it simultaneously introduces a new set of security vulnerabilities by distributing the attack surface. Physically exposed \textbf{edge devices (D1)} are susceptible to tampering, theft, and side-channel attacks, threatening the integrity of both the AI models and the sensitive data they process \cite{shafee2025,shafee2024}.

The distributed nature of Edge AI complicates centralized security management. Ensuring trust across a heterogeneous fleet of devices—from microcontrollers to edge servers—requires robust mechanisms like secure boot, trusted execution environments (TEEs), and homomorphic encryption. Furthermore, paradigms like \textbf{Federated Learning (D2)} are being adopted as a privacy-preserving training method, but they themselves introduce new challenges related to the security of aggregated model updates \cite{villar2023,jayanth2024}. A comprehensive, standardized security framework for these heterogeneous ecosystems is still nascent, representing a critical gap for sensitive \textbf{Application Domains (D3)} like healthcare and industrial control.

\subsection{Model Management and Deployment: The Operational Bottleneck}
The lifecycle management of AI models across vast, geographically dispersed edge deployments presents a profound operational challenge. The core of this problem is the extreme \textbf{hardware and software heterogeneity (D4)} across the edge continuum. Deploying, updating, and maintaining models on millions of devices, each with potentially different capabilities, requires sophisticated orchestration platforms that are both robust and secure \cite{choudhary2025,anchitaalagammai2025}.

Over-the-air (OTA) updates must be efficient and fault-tolerant, especially for devices with intermittent connectivity. Strategies such as A/B partitioning and rollback capabilities are essential to ensure reliability. Furthermore, maintaining model consistency and version control across this diverse fleet, while simultaneously debugging issues in the field, adds significant complexity to the DevOps cycle for Edge AI, potentially stalling the deployment of new applications \cite{baresi2019}.

\subsection{Power Consumption: The Energy Efficiency Frontier}
\label{subsec:power}

Energy efficiency represents arguably the paramount challenge for battery-operated or energy-harvesting devices operating at the \textbf{Device Edge (D1)}. The fundamental tension between the objective of continuous, always-on sensing and inference capabilities and the requirement for multi-year operational lifespans necessitates innovative solutions. Although specialized low-power \textbf{accelerators (D4)} provide partial mitigation, a comprehensive, system-wide approach to power management remains essential, requiring optimization across every system component from sensors to communication modules \cite{soro2021,lee2021}.

Current research converges on three synergistic strategies to advance the energy-efficiency frontier. First, \textbf{hardware-level optimization} employs quantized models—utilizing 8-bit integers rather than floating-point representations—to achieve 3–5× reductions in memory bandwidth and computational energy consumption. This approach is complemented by dedicated low-power cores that handle simple always-on tasks at microwatt power levels, thereby maintaining main processors in deep sleep states for extended durations \cite{husom2025,soro2021}. Second, \textbf{algorithmic efficiency} techniques, including pruning and sparsity exploitation, eliminate redundant operations to reduce energy consumption by 30–60\%. Event-triggered inference further enhances efficiency by radically reducing the duty cycle, processing data exclusively in response to meaningful sensor events such as detected motion \cite{muhoza2023,lee2021}. Third, \textbf{system-wide power gating }implements techniques including dynamic voltage and frequency scaling (DVFS) to adjust computational resources in real-time, while selective peripheral disabling powers down unused sensors and radios between inference cycles \cite{meng2023,katare2025}.

Significant challenges persist, particularly regarding the minimization of accuracy penalties associated with extreme quantization and the development of standardized power management interfaces capable of operating across heterogeneous \textbf{hardware platforms (D4)}. The most promising trajectory forward appears to lie in hybrid approaches that combine hardware-software co-design with adaptive, learning-based algorithms \cite{lee2021,meng2023}.

\subsection{Connectivity: The Reliability Bottleneck}
The efficacy of distributed Edge AI systems is fundamentally dependent on reliable and sufficient network connectivity, encompassing both bandwidth and reliability \cite{mwase2022}. Unlike cloud computing, the edge continuum often operates in environments with unstable or low-bandwidth connections, such as rural areas, moving vehicles, or dense industrial settings. This variability directly complicates model management (e.g., failed OTA updates) and constrains the feasibility of collaborative paradigms like Federated Learning, which require frequent model update exchanges. Furthermore, strategies to overcome poor connectivity, such as data buffering or more complex compression algorithms, can exacerbate power consumption challenges, creating a critical trade-off between operational reliability and energy efficiency.

\subsection{Interoperability and Standardization: The Integration Challenge}

The fragmented nature of the Edge AI ecosystem presents a critical barrier to widespread adoption. The proliferation of proprietary hardware accelerators (D4), diverse software frameworks, and incompatible communication protocols frequently results in vendor lock-in, increased development complexity, and substantial challenges in integrating disparate components into cohesive, scalable systems \cite{dave2024,stanko2024}.

This absence of common standards significantly impedes innovation and elevates development costs. While industry consortia and open-source initiatives are actively working to establish common APIs—such as the Open Neural Network Exchange (ONNX)—standardized data formats, and consistent deployment methodologies, the remarkable diversity of the edge landscape renders widespread adoption a long-term objective \cite{letaief2022,zhou2019}. Achieving genuine interoperability remains crucial for unlocking the full potential of Edge AI technologies and enabling seamless collaboration between devices operating across different \textbf{Deployment Locations (D1)}, as required by many advanced\textbf{ Applications (D3)}.

\subsection{Synthesis of Challenges}
The challenges facing Edge AI are not isolated but are deeply interconnected. The drive for greater energy efficiency (\textbf{Power}) can exacerbate security vulnerabilities. Unreliable \textbf{Connectivity} complicates model management and can constrain processing capabilities. The resource constraints of a chosen \textbf{Hardware (D4)} platform directly limit the complexity of models that can be deployed, affecting the achievable \textbf{Processing Capability (D2)} and thus the feasible \textbf{Applications (D3)}. Navigating this complex web of trade-offs is the central task for Edge AI system architects, requiring a holistic approach informed by the multi-dimensional perspective provided by our taxonomic framework.

\section{Future Research Horizons and Emerging Paradigms}
\label{sec:future}

\begin{figure}[htbp]
\centering
\includegraphics[width=\textwidth]{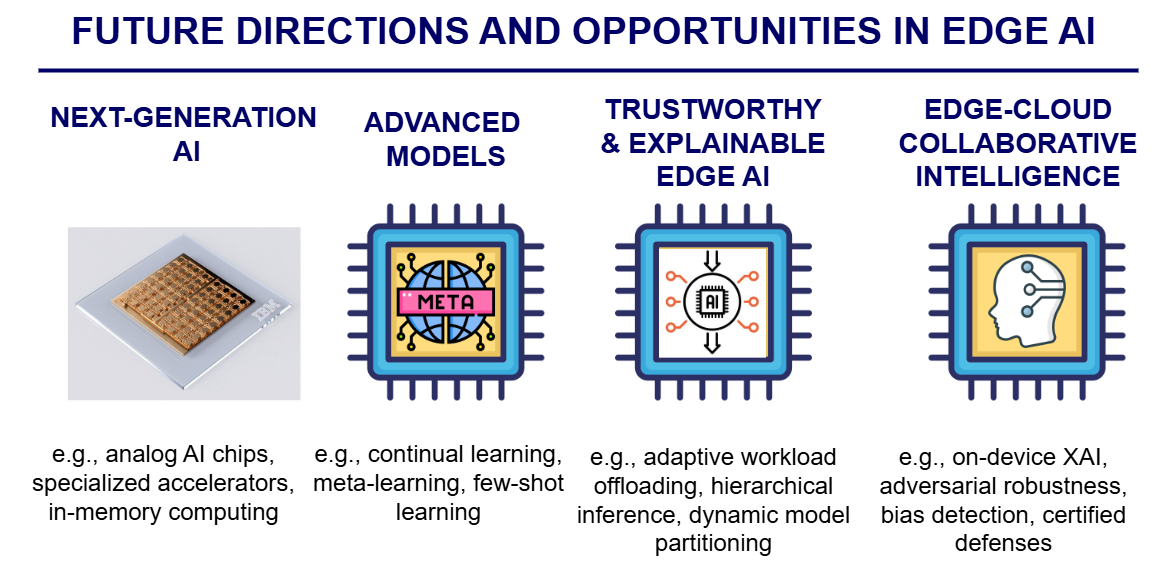}
\caption{Emerging research vectors in Edge AI, spanning next-generation hardware paradigms, advanced algorithmic capabilities, and systemic architectures for collaborative intelligence.}
\label{fig:future}
\end{figure}

The evolution of Edge AI is accelerating, driven by the imperative to overcome current limitations and unlock new frontiers of decentralized intelligence. The future landscape will be characterized by unprecedented hardware specialization, algorithms capable of autonomous adaptation, and the seamless, trust-aware orchestration of intelligence across the edge-to-cloud continuum. This section projects these future trajectories, framing them as natural progressions within the multi-dimensional taxonomy that structures this review. These emerging opportunities, summarized in Figure \ref{fig:future}, promise to address the challenges outlined in Section 5 and radically expand the application horizon of Edge AI.

\subsection{Next-Generation Hardware: Redefining the Performance-Energy Frontier (D4)}

The relentless pursuit of enhanced energy efficiency and computational density is driving a paradigm shift within the \textbf{Hardware Type (D4) }dimension, heralding a transition from incremental improvements to fundamentally novel computing architectures. Future Edge AI hardware development will be characterized by four transformative trajectories that collectively redefine the performance-energy frontier.

First, the emergence of \textbf{hyper-specialized accelerators} marks a departure from general-purpose AI chips toward processors meticulously optimized for specific model architectures—including transformers and graph neural networks—and specialized data modalities such as sparse and event-based data. This architectural evolution promises order-of-magnitude efficiency gains for niche Application Domains (D3), particularly in real-time sensor fusion and edge-based natural language understanding \cite{li2020,haris2024}.

Second, \textbf{in-memory and near-memory computing (IMC/NMC)} architectures represent a fundamental rethinking of computational paradigms by executing operations directly within or adjacent to memory cells through emerging technologies including memristors and magnetoresistive random-access memory (MRAM). This approach substantially mitigates the von Neumann bottleneck, dramatically reducing data movement energy and enabling unprecedented ultra-low-power, high-throughput inference capabilities essential for data-intensive applications at the Device Edge (D1) \cite{khwa2025,wen2024}.

Third,\textbf{ post-digital computing paradigms} are transitioning from theoretical constructs to practical implementations, with analog AI and optical computing offering transformative potential. Analog compute-in-memory techniques, leveraging fundamental physical principles including Ohm's and Kirchhoff's laws for matrix operations, demonstrate potential for 10–100× energy reduction in always-on processing scenarios, particularly for sparse data applications in industrial monitoring and environmental sensing \cite{fick2022,pile2024}.

Finally, the development of \textbf{self-powered intelligent systems} embodies the convergence of advanced energy harvesting technologies—spanning solar, kinetic, and radio frequency harvesting—with ultra-low-power AI processors. This integration enables perpetually operational, maintenance-free devices that fundamentally expand viable Deployment Locations (D1) to include remote, inaccessible, and hazardous environments, thereby transcending conventional power constraints in intelligent sensing applications \cite{trivedi2025,ammar2022}.

Collectively, these innovations represent not merely incremental advances but rather a fundamental reimagining of computational approaches that will enable previously impossible Edge AI applications while addressing critical energy constraints.

\subsection{Advanced Algorithms: Towards Adaptive and Resource-Aware Intelligence (D2)}

Algorithmic research is poised to fundamentally transform Edge AI systems by imbuing them with greater autonomy, efficiency, and contextual understanding, thereby substantially advancing the capabilities within the \textbf{Processing Capability (D2)} dimension. This evolution will manifest through several critical research directions that collectively address the unique constraints and opportunities of edge computing environments.

A primary focus will be the development of \textbf{continual and lifelong learning systems}, which will progressively replace static, pre-deployed models with architectures capable of incremental learning from continuous data streams. Such systems will enable models to adapt to concept drift and personalize to local environments without suffering catastrophic forgetting, thereby drastically reducing the need for costly retraining and redeployment cycles \cite{soltoggio2024,wang2025b}.

Concurrently, \textbf{meta-learning and few-shot learning} approaches will emerge as crucial enablers for agile deployment in diverse and dynamic scenarios where labeled data is inherently scarce. These techniques will facilitate "plug-and-play" intelligence, allowing pre-optimized models to rapidly specialize for new sensors or environmental conditions at the edge \cite{chen2024,lu2024}.

The deployment of \textbf{Tiny Reinforcement Learning (TinyRL) agents} directly on edge devices will represent another significant advancement, enabling autonomous, real-time decision-making and control loops. This capability proves particularly transformative for Application Domains (D3) such as robotics, industrial automation, and network management, where systems must learn and react to complex environments without the latency of cloud dependency \cite{wu2024,du2023}. 

Finally, as Edge AI penetrates increasingly critical applications, research will focus on developing \textbf{explainable AI (XAI)} techniques specifically designed for \textbf{resource-constrained environments}. These methods will provide interpretable rationales for model decisions without overwhelming the limited computational and memory resources of edge hardware, thereby addressing growing demands for transparency and accountability in autonomous systems \cite{hassanien2023,nguyen2025}.

Together, these algorithmic advancements will create a new generation of adaptive, resource-aware intelligence systems capable of operating effectively within the stringent constraints of edge computing environments while maintaining sophisticated learning and decision-making capabilities.

\subsection{Edge-Cloud Collaborative Intelligence: The Emergence of a Cognitive Continuum}

The conventional rigid dichotomy between edge and cloud computing is evolving toward a fluid, hierarchical intelligence continuum, representing the maturation of the \textbf{Deployment Location (D1)} dimension into a dynamic, integrated system. This architectural shift will be characterized by several defining features that collectively enable more efficient and responsive intelligent systems.

Future architectures will be defined by \textbf{adaptive hierarchical systems} that dynamically distribute intelligence across computational tiers. In this paradigm, TinyML models will perform initial filtering and time-critical inference at the Device Edge (D1), while more complex analytical tasks—such as multi-sensor fusion and long-term trend analysis—will be offloaded to Network Edge servers. The cloud will increasingly specialize in large-scale model training and global aggregation, thereby establishing a seamless flow of computation and data across the entire continuum \cite{dhakad2024,firouzi2022}.

Complementing this architectural evolution, \textbf{intelligent dynamic offloading} mechanisms will employ AI-powered controllers to make real-time decisions about workload placement based on a comprehensive assessment of network conditions, computational load, energy availability, and application requirements. This approach will systematically optimize the complex trade-offs between latency, bandwidth, accuracy, and power consumption across the entire system \cite{kumari2024,li2023}.

Furthermore, \textbf{enhanced federated learning frameworks} will advance beyond simple update averaging to incorporate stronger privacy guarantees—including differential privacy and homomorphic encryption—along with robust aggregation algorithms capable of handling extreme non-IID data and device heterogeneity. These frameworks will also integrate mechanisms for detecting and mitigating malicious participants, thereby creating a truly scalable and secure solution for privacy-sensitive domains \cite{rane2024,wang2023}.

\begin{figure}[htbp]
\centering
\includegraphics[width=0.7\textwidth]{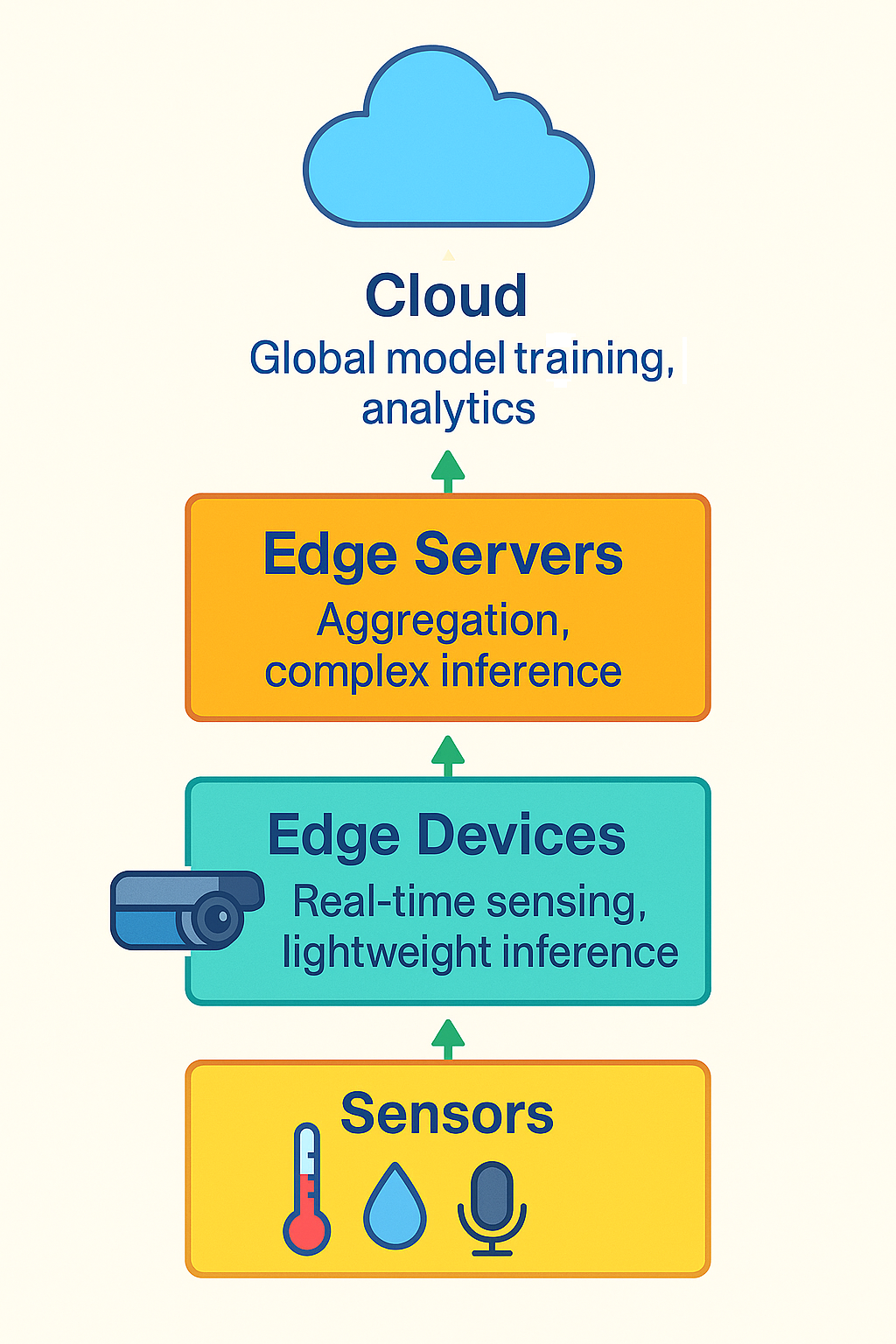}
\caption{The future cognitive continuum: A dynamic, hierarchical architecture that intelligently partitions and orchestrates AI workloads across the device-edge-cloud spectrum based on real-time constraints and requirements.}
\label{fig:hierarchy}
\end{figure}

This collaborative paradigm, illustrated in Figure \ref{fig:hierarchy}, envisions a future where embedded TinyDL models in smart cameras perform real-time object detection, edge servers aggregate multiple feeds for sophisticated crowd analytics, and cloud-based federated learning systems securely improve global models—all functioning as a single, cohesive intelligent system that dynamically adapts to changing conditions and requirements.

\subsection{Trustworthy and Explainable Edge AI: The Foundation for Adoption}

For Edge AI systems to achieve deployment in truly critical applications, trust must be elevated to a first-class design constraint, systematically integrated across both the\textbf{ Processing Capability (D2)} and \textbf{Application Domain (D3)} dimensions. This imperative will drive several critical research thrusts that collectively address the unique challenges of trustworthy computing in resource-constrained environments.

A primary research focus will center on developing \textbf{robustness against adversarial attacks} through defense mechanisms specifically designed for the computational constraints of edge hardware. This includes advancing efficient adversarial training techniques, input purification methods, and runtime detection algorithms that can secure Edge AI systems against evolving threat landscapes without compromising operational efficiency \cite{zhong2024,yi2023}.

Equally important will be addressing the unique challenges of \textbf{bias detection and mitigation} in models trained on decentralized, non-IID edge data. Research must develop specialized techniques for detecting, quantifying, and mitigating bias directly on edge devices or during federated learning processes, ensuring equitable and ethical AI outcomes across diverse deployment scenarios \cite{hutiri2023,katare2022}.

Finally, achieving practical \textbf{on-device explainability} will require methods capable of generating concise, meaningful explanations for model decisions directly on edge hardware, utilizing only a fraction of the resources required for inference itself. This capability will prove paramount for establishing user trust, ensuring regulatory compliance, and enabling effective developer debugging in field deployments \cite{xu2024,xu2024b}.

Together, these research directions will establish the foundational trustworthiness necessary for Edge AI systems to expand into increasingly critical applications while maintaining the efficiency requirements inherent to edge computing environments.

\subsection{The Role of 6G and Advanced Networking: The Connective Tissue}

Next-generation wireless networks, particularly 6G, will transcend their conventional role as mere data conduits to become the intelligent connective tissue that unifies the edge computing continuum, thereby directly enabling transformative A\textbf{pplication Domains (D3)}. This evolution will manifest through several foundational capabilities that collectively redefine the relationship between networking and distributed intelligence.

6G networks are anticipated to provide \textbf{sub-millisecond latency and holographic-type communication capabilities}, which will fundamentally unlock applications requiring unprecedented responsiveness. These advancements will enable collaborative swarm robotics, autonomous vehicle platooning, and immersive extended reality (XR) experiences where haptic feedback and real-time interaction demand near-instantaneous communication \cite{zhang2024,zhou2025}.

Furthermore, the \textbf{deep integration of artificial intelligence and communication} technologies will transform the network infrastructure itself into an inherently intelligent system. Through the embedding of AI capabilities directly within the radio access network (RAN), future networks will dynamically allocate resources, predict network states, and optimize performance parameters specifically tailored to Edge AI workload requirements \cite{zhang2024b,barker2025}.

Additionally, the emergence of \textbf{sensing-as-a-service} capabilities will represent a paradigm shift in network functionality. 6G infrastructure is projected to incorporate integrated sensing technologies, effectively transforming the network into a distributed sensor array that can provide rich perceptual context to Edge AI devices, thereby significantly enhancing their situational awareness and operational capabilities across diverse environments.

Collectively, these advancements will establish next-generation networks as active enablers of Edge AI systems rather than passive communication channels, creating a symbiotic relationship between networking capabilities and distributed intelligence that will drive innovation across the entire computing continuum.

\subsection{Synthesis: Towards a Cognitive Edge}
The convergence of these directions points to a future of a "Cognitive Edge"—an intelligent, self-adapting, and trustworthy fabric of distributed computation. This evolution will be characterized by a shift from deploying static models on isolated devices to orchestrating dynamic intelligence across a continuum of heterogeneous resources. The multi-dimensional taxonomy presented in this review provides the essential framework for navigating this complex and exciting future, outlining the inter-dependencies between hardware, algorithms, deployment strategies, and applications that will define the next decade of Edge AI innovation.



\section{Conclusion}
\label{sec:conclusion}
Edge Artificial Intelligence represents a paradigm shift in computational architecture, fundamentally redefining how intelligent systems are designed, deployed, and integrated into the physical world. This comprehensive review has systematically charted the evolution, current state, and future trajectory of Edge AI, offering a holistic analysis through a novel multi-dimensional taxonomy that integrates deployment location, processing capability, application domain, and hardware type.

Our analysis reveals that the journey from cloud-centric computing to distributed edge intelligence has been neither accidental nor instantaneous. It is the result of a coherent evolution through foundational technologies—CDNs, fog computing, and mobile edge computing—each solving critical limitations of its predecessor and collectively paving the way for modern paradigms like TinyML, TinyDL, TinyRL, and federated learning. This historical contextualization, often neglected in prior surveys, provides essential perspective for understanding current developments and future directions.

Through our systematic methodology and taxonomic framework, we have demonstrated that the contemporary Edge AI landscape is characterized by sophisticated hardware-software co-design across a spectrum of resource constraints. We have further shown how these technological capabilities enable transformative applications across diverse sectors—from healthcare and industrial automation to autonomous systems and smart cities—each with unique requirements for latency, privacy, and autonomy.

However, this review also identifies significant challenges that constrain widespread adoption. Resource constraints, security vulnerabilities, model management complexities, and power consumption limitations present formidable hurdles that require continued innovation. These challenges are not isolated but interconnected, demanding holistic solutions that address trade-offs across hardware, software, and deployment architectures.

Looking forward, we project that Edge AI's future lies in several key directions: (1) next-generation hardware paradigms that redefine energy-performance trade-offs through in-memory computing, analog AI, and specialized accelerators; (2) advanced algorithms capable of continuous adaptation, few-shot learning, and explainable decision-making within resource constraints; (3) seamless edge-cloud collaborative intelligence that dynamically distributes workloads across the computational continuum; and (4) the integration of trustworthiness and explainability as fundamental design principles rather than afterthoughts.

The realization of Edge AI's full potential will require unprecedented interdisciplinary collaboration across hardware engineering, computer systems, algorithm design, and application domains. As 5G/6G networks mature and AI workloads become increasingly pervasive, the principles and architectures discussed in this review will become central to next-generation intelligent systems.

Ultimately, Edge AI is poised to create a future where artificial intelligence becomes truly ubiquitous—embedded not just in devices but woven into the very fabric of our environment, enabling responsive, intelligent, and autonomous systems that operate seamlessly within our physical world while respecting the constraints of resources, privacy, and energy. This survey provides a comprehensive foundation for researchers, practitioners, and policymakers to navigate and contribute to this rapidly evolving field.

\appendix
\section{Complete Edge AI Reference Taxonomy}

\begin{landscape}
\begin{longtable}
{@{}p{0.8cm}p{2.2cm}p{2.8cm}p{2cm}p{7.2cm}@{}}
\caption[Edge AI Reference Taxonomy]{Systematic classification of seminal Edge AI literature (2017--2025) organized by the review's taxonomic categories: hardware accelerators, TinyML/TinyDL/TinyRL paradigms, federated learning, edge systems (Fog/MEC/CDN), application domains, and emerging challenges.}\label{tab:edgeai_refs}\\

\toprule
\textbf{Year} & \textbf{Category} & \textbf{Subcategory} & \textbf{Reference} & \textbf{Key Contribution} \\
\midrule
\endfirsthead
\toprule
\textbf{Year} & \textbf{Category} & \textbf{Subcategory} & \textbf{Reference} & \textbf{Key Contribution} \\
\midrule
\endhead
\hline
\multicolumn{5}{r@{}}{\footnotesize\itshape Continued on next page} \\
\endfoot
\bottomrule
\endlastfoot

2020 & \textbf{Hardware} & \textbf{ASIC} & \cite{liang2024} & Google Edge TPU architecture analysis \\
2021 & \textbf{Hardware} & \textbf{GPU} & \cite{bouzidi2022} & NVIDIA Jetson performance profiling \\
2022 & \textbf{Hardware} & \textbf{FPGA} & \cite{liu2024} & Real-time SRAM-based acceleration \\
2023 & \textbf{Hardware} & \textbf{Neuromorphic} & \cite{das2024} & Intel Loihi2 edge deployment \\
2024 & \textbf{Hardware} & \textbf{CiM} & \cite{khwa2025} & Memristor-based compute-in-memory \\
2024 & \textbf{Hardware} & \textbf{Survey} & \cite{alam2024} & Comparative analysis of 32 accelerators \\
2025 & \textbf{Hardware} & \textbf{Analog} & \cite{fick2022} & Mythic analog AI chip case study \\

\specialrule{0.1pt}{3pt}{3pt}
2017 & \textbf{TinyML} & \textbf{DL} & \cite{darvish2017} & First just-in-time DL compilation \\
2020 & \textbf{TinyML} & \textbf{Tools} & \cite{ray2022} & TensorFlow Lite Micro framework \\
2022 & \textbf{TinyML} & \textbf{MCU} & \cite{immonen2022} & ARM Cortex-M4 optimizations \\
2023 & \textbf{TinyML} & \textbf{Vision} & \cite{burrello2024} & Neural architecture search for MCUs \\
2024 & \textbf{TinyML} & \textbf{Survey} & \cite{lin2023} & State-of-the-art techniques review \\

\specialrule{0.1pt}{3pt}{3pt}
2023 & \textbf{TinyDL} & \textbf{Survey} & \cite{somvanshi2025} & From TinyML to TinyDL: A comprehensive survey \\
2023 & \textbf{TinyDL} & \textbf{Architecture} & \cite{roveri2023} & Analysis of deep learning on microcontrollers \\
2024 & \textbf{TinyDL} & \textbf{Optimization} & \cite{burrello2024} & Hardware-aware NAS for TinyDL models \\

\specialrule{0.1pt}{3pt}{3pt}
2024 & \textbf{TinyRL} & \textbf{Algorithms} & \cite{wu2024} & Design principles for RL on edge devices \\
2023 & \textbf{TinyRL} & \textbf{Applications} & \cite{du2023} & Diffusion-based RL for edge-generated content \\

\specialrule{0.1pt}{3pt}{3pt}
2019 & \textbf{Federated} & \textbf{Foundational} & \cite{wang2019} & First edge FL framework \\
2021 & \textbf{Federated} & \textbf{Privacy} & \cite{brecko2022} & Differential privacy enhancements \\
2022 & \textbf{Federated} & \textbf{Survey} & \cite{abreha2022} & Analysis of 58 deployments \\
2023 & \textbf{Federated} & \textbf{Robustness} & \cite{villar2023} & Adversarial attack defenses \\

\specialrule{0.1pt}{3pt}{3pt}
2018 & \textbf{Edge} & \textbf{Fog} & \cite{gupta2018} & Fog computing architecture \\
2020 & \textbf{Edge} & \textbf{MEC} & \cite{ahmed2024} & 5G MEC standardization \\
2021 & \textbf{Edge} & \textbf{CDN} & \cite{zolfaghari2021} & AI-enhanced content delivery \\
2022 & \textbf{Edge} & \textbf{Survey} & \cite{srirama2023} & 10-year evolution analysis \\
2024 & \textbf{Edge} & \textbf{MEC} & \cite{khan2022} & MEC for video streaming and VR \\

\specialrule{0.1pt}{3pt}{3pt}
2020 & \textbf{Apps} & \textbf{Healthcare} & \cite{rocha2024} & Wearable ECG monitoring \\
2021 & \textbf{Apps} & \textbf{Automotive} & \cite{xie2024} & Real-time object detection \\
2022 & \textbf{Apps} & \textbf{Industrial IoT} & \cite{artiushenko2024} & Resource-efficient Edge AI for predictive maintenance \\
2022 & \textbf{Apps} & \textbf{Industry} & \cite{artiushenko2024} & Predictive maintenance systems \\

\specialrule{0.1pt}{3pt}{3pt}
2019 & \textbf{Challenges} & \textbf{Privacy} & \cite{Sun2020} & Edge data protection framework \\
2021 & \textbf{Challenges} & \textbf{Power} & \cite{soro2021} & Energy harvesting techniques \\
2022 & \textbf{Challenges} & \textbf{Security} & \cite{shafee2025} & Attack vector taxonomy \\

\specialrule{0.1pt}{3pt}{3pt}
2022 & \textbf{Future} & \textbf{Continual} & \cite{soltoggio2024} & Lifelong learning algorithms \\
2023 & \textbf{Future} & \textbf{Bio} & \cite{trivedi2025} & Neuromorphic edge systems \\
2025 & \textbf{Future} & \textbf{6G} & \cite{zhou2025} & AI-optimized RAN architectures \\
\end{longtable}
\end{landscape}




\bibliographystyle{unsrt}

\end{document}